\documentclass[10pt,twocolumn]{article}
\usepackage[utf8]{inputenc} 
\usepackage{graphicx} 
\usepackage{amsmath} 
\usepackage{hyperref} 
\usepackage{geometry} 
\usepackage{titlesec} 

\titleformat{\section}{\large\bfseries}{\thesection}{1em}{}

\title{\textbf{Adaptive Few-Shot Learning (AFSL): Tackling Data Scarcity with Stability, Robustness, and Versatility}}
\author{\textbf{Rishabh Agrawal} \\ University of Illinois at Urbana-Champaign}

\date{} 
\begin{document}

\maketitle

\begin{abstract}
Few-shot learning (FSL) has emerged as a groundbreaking machine learning paradigm, enabling models to generalize effectively with minimal labeled data. This approach is particularly vital in domains where data collection and annotation are resource-intensive, such as healthcare diagnostics, robotics, and natural language processing. Despite its transformative potential, FSL is challenged by issues such as sensitivity to initialization, difficulty in adapting to heterogeneous domains, and vulnerability to noisy or mislabeled datasets. This paper introduces \textit{Adaptive Few-Shot Learning (AFSL)}, a novel and comprehensive framework that addresses these challenges by integrating advancements in meta-learning, domain alignment, noise resilience, and multi-modal adaptability. AFSL consists of four key modules: a Dynamic Stability Module to enhance performance consistency, a Contextual Domain Alignment Module to handle domain shifts, a Noise-Adaptive Resilience Module to mitigate the impact of noisy data, and a Multi-Modal Fusion Module to enable seamless integration of diverse data modalities. This paper also explores strategies such as task-aware data augmentation, semi-supervised learning, and explainable AI techniques to further improve FSL’s applicability and robustness. By addressing these critical challenges, AFSL demonstrates its capability to deliver scalable, reliable, and impactful solutions across various high-stakes domains.
\end{abstract}

\subsection*{Keywords}
Few-Shot Learning (FSL), Adaptive Few-Shot Learning (AFSL), Meta-Learning, Domain Adaptation, Robust Learning, Data Augmentation, Semi-Supervised Learning, Multi-Modal Learning, Noise-Resilient Learning, Adversarial Learning, Model Generalization.

\section{Introduction}

Few-shot learning (FSL) has emerged as a transformative approach in machine learning, addressing the significant challenge of training models with minimal labeled data. Unlike traditional paradigms that depend on large-scale annotated datasets, FSL allows models to generalize effectively from only a few examples. This capability is particularly crucial in domains where data collection and annotation are expensive or infeasible, such as rare disease diagnosis in healthcare, language processing for low-resource languages, and robotic task adaptation in dynamic environments.

Despite its potential, FSL faces several critical challenges that limit its scalability and adoption in real-world applications. A key issue is \textbf{sensitivity to randomness}, where model performance is highly variable due to random initialization or sampling strategies. This inconsistency complicates reproducibility and hinders the development of reliable benchmarks. Another major challenge is \textbf{domain adaptation}, as FSL models often fail to generalize effectively when source and target domains have significant differences in data distributions or class representations. This limitation is particularly problematic in scenarios such as medical imaging, where population-level variations in data can drastically affect model performance. Additionally, \textbf{robustness to noisy data} remains a significant obstacle, with real-world datasets often contaminated by mislabeled or outlier examples that degrade model accuracy, especially in high-stakes domains like finance and autonomous systems.

To address these challenges holistically, this paper introduces the \textit{Adaptive Few-Shot Learning (AFSL)} framework. AFSL integrates advancements in meta-learning, domain alignment, noise resilience, and multi-modal learning to create a unified, modular architecture. Its \textbf{Dynamic Stability Module} mitigates sensitivity to randomness by leveraging ensemble-based meta-learning and variance-reduction techniques, ensuring consistent performance across tasks. The \textbf{Contextual Domain Alignment Module} addresses domain shifts using adversarial learning and hierarchical alignment strategies to align source and target domains effectively. To enhance robustness, the \textbf{Noise-Adaptive Resilience Module} dynamically suppresses noisy or mislabeled data through attention-based weighting and noise-aware loss functions. Finally, the \textbf{Multi-Modal Fusion Module} extends FSL’s capabilities to integrate diverse data modalities, enabling seamless adaptation to complex multi-modal tasks.

By combining these modules, AFSL provides a comprehensive solution to the limitations of current FSL approaches. This paper explores the state-of-the-art methodologies, identifies gaps in existing research, and demonstrates how AFSL achieves significant improvements in generalization, stability, and robustness. The proposed framework sets a new benchmark for Few-Shot Learning, unlocking its potential across critical domains such as healthcare, robotics, natural language processing, and beyond.

\section{Related Work}

Few-shot learning (FSL) has garnered significant attention as a promising approach to address the challenges posed by data scarcity. This section reviews key advancements in meta-learning, domain adaptation, noise resilience, and semi-supervised learning, highlighting existing limitations and motivating the need for a novel framework.

\subsection{Meta-Learning}
Meta-learning frameworks, such as Model-Agnostic Meta-Learning (MAML) and Prototypical Networks, have demonstrated significant success in enabling models to generalize across tasks with minimal labeled data. MAML optimizes for initial model parameters that can adapt efficiently to new tasks with minimal updates, while Prototypical Networks classify based on learned class prototypes in an embedding space. Despite their strengths, these methods exhibit sensitivity to initialization and sampling strategies, resulting in instability and performance variability. Recent advancements, such as task-specific auxiliary objectives and hybrid models combining supervised and self-supervised learning, have partially addressed these issues but remain constrained by high computational costs and limited scalability.

\subsection{Domain Adaptation}
Domain adaptation techniques, including Transfer Learning and Domain-Adversarial Neural Networks (DANNs), aim to align feature distributions between source and target domains. Transfer Learning adapts pre-trained models to target tasks, whereas DANNs use adversarial learning to achieve domain invariance. However, these approaches struggle with extreme domain shifts, such as non-overlapping classes or highly heterogeneous feature distributions. Hierarchical alignment strategies and adversarial domain mapping have emerged as promising solutions, but their integration into few-shot scenarios remains underexplored.

\subsection{Noise Resilience}
The presence of noisy and mislabeled data in real-world scenarios poses significant challenges for FSL models. Techniques such as Robust Attentive Profile Networks (RapNets) and noise-filtering mechanisms dynamically adjust the weights of training samples to mitigate the impact of unreliable data. While these methods improve robustness, they often introduce additional computational overhead, limiting their practicality for resource-constrained applications. Advances in noise-aware attention mechanisms and consistency regularization have shown potential for addressing these limitations but require further optimization.

\subsection{Semi-Supervised Learning}
Semi-supervised learning approaches have been instrumental in leveraging unlabeled data to enhance FSL performance. Pseudo-labeling and consistency regularization are widely adopted, enabling models to utilize large volumes of unlabeled data by assigning labels to samples based on confident predictions. However, the quality of pseudo-labels significantly affects performance, and errors can propagate during training. Unsupervised meta-learning, which constructs tasks directly from unlabeled datasets, offers an alternative by identifying clusters and patterns without requiring labeled data. Recent innovations have combined semi-supervised and unsupervised learning techniques, demonstrating improved generalization on benchmark datasets.

\begin{figure}[h]
    \centering
    \includegraphics[width=0.45\textwidth]{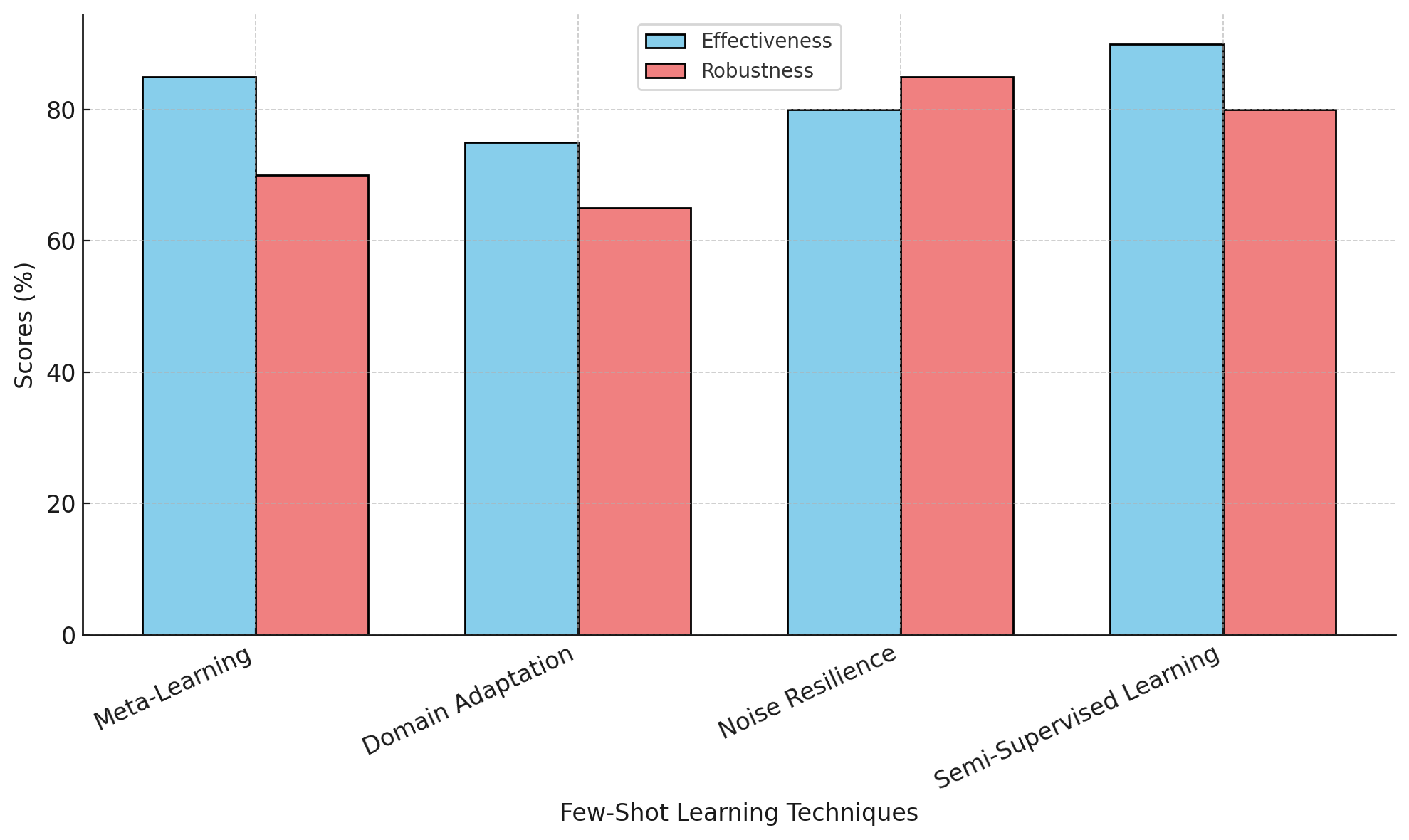} 
    \caption{Comparison of Few-Shot Learning Techniques across Key Dimensions.}
    \label{fig:related_work}
\end{figure}

\section{Proposed Framework: Adaptive Few-Shot Learning (AFSL)}

To address the persistent challenges in Few-Shot Learning (FSL), we propose \textit{Adaptive Few-Shot Learning (AFSL)}, a comprehensive framework designed to integrate advancements in meta-learning, domain generalization, noise resilience, and multi-modal adaptability. The framework introduces four interconnected modules, each addressing a specific bottleneck to ensure robust, scalable, and adaptable solutions for real-world FSL scenarios.

\begin{figure}[h]
    \centering
    \includegraphics[width=0.45\textwidth]{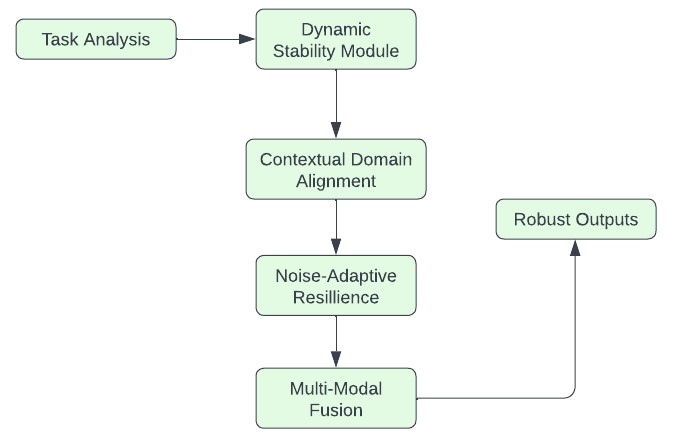} 
    \caption{Workflow of the Adaptive Few-Shot Learning (AFSL) Framework.}
    \label{fig:related_work}
\end{figure}

\subsection{Dynamic Stability Module}
FSL models are often sensitive to randomness in initialization and sampling, resulting in inconsistent performance. The Dynamic Stability Module mitigates this issue through \textbf{dynamic ensemble optimization}, where meta-models are dynamically selected and weighted based on task-specific complexity. By leveraging task embeddings, this module identifies task similarities and adapts ensemble strategies in real time. Additionally, it incorporates \textbf{gradient noise reduction} during meta-training to minimize fluctuations, ensuring reproducible and stable results across diverse tasks.

\subsection{Contextual Domain Alignment Module}
Domain shifts, where the source and target domains differ significantly in feature distributions, remain a major hurdle in FSL. The Contextual Domain Alignment Module addresses this challenge using \textbf{adversarial alignment} combined with \textbf{contextual meta-learning}. Domain-specific features are dynamically identified and aligned using hierarchical feature alignment, preserving both global and local task-specific information. Furthermore, \textbf{contrastive domain embeddings} are used to ensure that features from distinct domains remain contextually aligned but separable, enabling effective adaptation across heterogeneous datasets.

\subsection{Noise-Adaptive Resilience Module}
Real-world datasets often contain mislabeled or noisy data, which degrades model performance. This module utilizes \textbf{attention-guided noise filtering} and \textbf{self-supervised consistency checks} to mitigate noise effects. Noise-Aware Attention Networks (NANets) dynamically assign weights to training samples based on reliability, filtering out noisy or mislabeled examples. A \textbf{dual-loss framework} combines a noise-aware loss function with consistency-based regularization, ensuring stable predictions across different data augmentations.

\subsection{Multi-Modal Fusion Module}
Modern applications increasingly involve multi-modal data, such as images, text, and audio. The Multi-Modal Fusion Module enables seamless integration of these diverse modalities using \textbf{cross-attention transformers}. By creating \textbf{shared embedding spaces}, this module aligns features across modalities, facilitating \textbf{cross-modal transfer learning}. For example, textual descriptions can enhance image classification by providing semantic context, while audio signals can improve scene understanding in robotics.

\section*{Integration and Workflow}
The AFSL framework seamlessly integrates its modules as follows:
\begin{enumerate}
    \item Tasks are first grouped by complexity, with the Dynamic Stability Module ensuring stable initialization and predictions.
    \item Source and target domain features are aligned using the Contextual Domain Alignment Module for effective transfer learning.
    \item Noisy and mislabeled data are filtered and corrected by the Noise-Adaptive Resilience Module.
    \item Multi-modal data is fused and aligned using the Multi-Modal Fusion Module, enabling robust learning across diverse modalities.
\end{enumerate}

\section*{Advantages of AFSL}
AFSL offers several key advantages:
\begin{itemize}
    \item \textbf{Robustness}: The dynamic stability and noise-adaptive mechanisms ensure consistent performance across datasets and tasks.
    \item \textbf{Scalability}: Multi-modal integration and domain alignment expand FSL’s applicability to complex, real-world scenarios.
    \item \textbf{Adaptability}: The modular architecture allows customization for specific domains, such as healthcare, robotics, and natural language processing.
\end{itemize}

By addressing the challenges of stability, domain alignment, noise resilience, and multi-modal learning, \textit{AFSL} sets a new benchmark for Few-Shot Learning, paving the way for scalable, reliable applications across diverse fields.

\section{Strategies for Improvement}

Few-Shot Learning (FSL) offers tremendous potential for addressing data scarcity challenges, but it is hindered by limitations in stability, adaptability, robustness to noise, and scalability. Recent advancements, including innovations in the Adaptive Few-Shot Learning (AFSL) framework, provide strategies for overcoming these limitations. This section explores improvements in meta-learning, data augmentation, semi-supervised learning, and multi-modal integration.

\begin{figure}[h]
    \centering
    \includegraphics[width=0.45\textwidth]{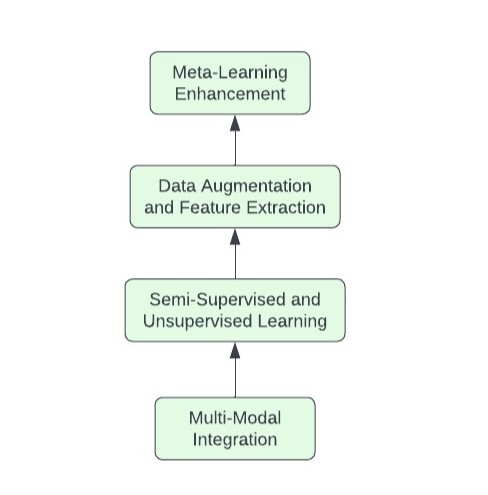} 
    \caption{Strategies for Improvement in Few-Shot Learning.}
    \label{fig:related_work}
\end{figure}

\subsection{Meta-Learning Enhancements}
Meta-learning, often referred to as “learning to learn,” is fundamental to FSL. A notable improvement involves \textbf{task decomposition}, where complex tasks are divided into manageable subtasks. Hierarchical task decomposition, as employed in AFSL’s Dynamic Stability Module, enables models to adapt better to diverse scenarios by identifying and leveraging task-specific similarities. 

\textbf{Auxiliary tasks with strong supervision} are another impactful enhancement. Tasks such as contrastive learning and masked token prediction strengthen feature representations by encouraging inter-class separability and intra-class cohesion. Recent transformer-based frameworks combining supervised and self-supervised objectives have demonstrated significant improvements on FSL benchmarks like miniImageNet. The integration of dynamic task embeddings in AFSL ensures these models achieve both stability and reproducibility.

\subsection{Data Augmentation and Feature Extraction}
Data augmentation is a cornerstone of FSL for mitigating data scarcity. Advanced techniques like \textbf{self-mixup} interpolate between data points and their labels, creating diverse synthetic samples that reduce overfitting and improve generalization. Studies in 2024 showed that self-mixup-augmented models outperformed traditional methods significantly on low-resource datasets.

\textbf{Task-aware data augmentation} further tailors augmentation strategies to specific tasks, ensuring augmented data aligns with the task’s unique characteristics. This approach, integrated into AFSL, minimizes noise and enhances model performance. 

In feature extraction, \textbf{calibration-adaptive downsampling} dynamically adjusts sampling rates based on task complexity, ensuring models prioritize high-quality features while suppressing noise. Transformer-based frameworks have shown the effectiveness of this technique in improving both accuracy and efficiency.

\subsection{Semi-Supervised and Unsupervised Learning}
Semi-supervised and unsupervised learning methods are critical for leveraging abundant unlabeled data. \textbf{Pseudo-labeling}, where a model iteratively assigns and refines labels for unlabeled data, has proven effective in expanding datasets and improving generalization. Semi-supervised frameworks integrating consistency regularization enforce stable predictions across augmented inputs, further enhancing model performance.

\textbf{Unsupervised meta-learning} constructs tasks directly from unlabeled datasets by identifying clusters or patterns. Clustering-based task generation, as explored in 2023, grouped similar samples into pseudo-classes, enabling effective task construction without relying on labeled data. This aligns closely with AFSL’s unsupervised task embeddings and pretraining modules.

\subsection{Multi-Modal Integration}
Modern FSL applications increasingly involve multi-modal data, such as images, text, and audio. AFSL’s Multi-Modal Fusion Module employs \textbf{cross-attention mechanisms} to align features across modalities, enabling robust cross-modal learning. For instance, integrating textual descriptions with images can enhance medical diagnosis, while combining audio and visual data improves robotics' scene understanding. By learning shared embedding spaces, FSL models generalize more effectively across diverse modalities, opening new possibilities for multi-modal applications in real-world scenarios.

\section{Applications of Few-Shot Learning}

Few-Shot Learning (FSL) has transformative potential across diverse domains, particularly in scenarios where labeled data is scarce or expensive to obtain. Its ability to generalize from minimal examples and adapt to new tasks makes it invaluable in real-world applications requiring rapid adaptability, scalability, and efficiency. Below, we explore key applications of FSL, highlighting advancements introduced by the Adaptive Few-Shot Learning (AFSL) framework.

\subsection{Healthcare}
FSL is critically important in healthcare, where labeled data is often limited, particularly for rare diseases and conditions. For example, FSL models trained on a small set of annotated medical images can generalize to classify new cases, reducing reliance on extensive datasets. AFSL’s multi-modal fusion module enhances medical diagnosis by integrating diverse data sources, such as patient histories (text) and medical imaging (visual data), resulting in more accurate and holistic predictions.

For diagnosing rare diseases like genetic disorders, AFSL’s noise-resilient mechanisms mitigate the impact of mislabeled data, ensuring reliable performance. Similarly, in drug discovery, FSL can analyze limited molecular datasets to predict potential drug interactions and identify novel drug candidates, accelerating research. Personalized medicine is another area where FSL shines, adapting treatment plans to an individual’s medical profile with minimal labeled data. In imaging tasks like tumor segmentation, AFSL’s task-aware data augmentation further enhances detection accuracy, supporting early diagnosis and treatment planning.

\subsection{Natural Language Processing (NLP)}
Few-Shot Learning has shown significant promise in natural language processing (NLP), particularly for applications like low-resource language translation and text classification. FSL enables pre-trained transformer models, fine-tuned using AFSL’s task decomposition techniques, to generalize efficiently to underrepresented languages. For example, indigenous language translation systems can be built using limited datasets, preserving endangered languages and improving accessibility.

Other NLP applications include conversational AI, where FSL identifies user intents with minimal examples, enhancing chatbot functionality. In domain-specific sentiment analysis, such as analyzing customer feedback, FSL is particularly effective. Furthermore, AFSL’s cross-modal transformers improve tasks like caption generation and multi-modal sentiment analysis by integrating textual and visual data.

\subsection{Robotics}
The adaptability of FSL is a game-changer in robotics, where tasks and environments vary significantly. With FSL, robots can quickly learn new tasks with minimal training examples, such as recognizing new tools in industrial automation. AFSL’s domain alignment module helps robots adapt to new environments, ensuring robust performance across diverse scenarios.

In unstructured settings like homes or disaster response areas, FSL-powered robots can learn skills such as assisting individuals with disabilities or navigating complex terrains by observing a few demonstrations. AFSL’s noise-resilient learning module ensures reliable performance even with ambiguous training data. One-shot imitation learning, where robots replicate tasks performed by humans, is another impactful application. By leveraging AFSL’s cross-modal capabilities, robots can integrate visual and auditory cues for improved task execution.

\begin{figure}[h]
    \centering
    \includegraphics[width=0.45\textwidth]{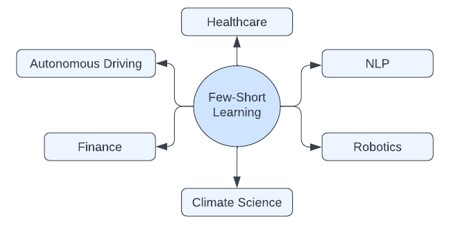} 
    \caption{Applications of Few-Shot Learning.}
    \label{fig:fig_5}
\end{figure}

\subsection{Other Emerging Domains}
Few-Shot Learning is expanding into fields such as climate science, finance, and autonomous driving, addressing unique challenges in each domain.

In climate science, FSL models rare events like hurricanes and wildfires using limited historical data. AFSL’s hierarchical alignment enables these models to adapt across regions with diverse climate patterns, improving disaster preparedness. 

In finance, FSL is revolutionizing fraud detection by identifying patterns in small datasets of fraudulent transactions. AFSL’s noise-resilient mechanisms ensure accurate anomaly detection while filtering noisy data. Additional applications include credit scoring and risk assessment, where labeled data is often sparse.

For autonomous driving, FSL allows self-driving vehicles to adapt to new environments and road conditions with minimal retraining. AFSL’s multi-modal fusion module integrates sensor inputs, visual data, and maps, enabling safer and more reliable decision-making.

\section{Future Directions}

Few-Shot Learning (FSL) offers immense potential for addressing challenges in data-scarce environments. Future research directions, building on the Adaptive Few-Shot Learning (AFSL) framework, focus on enhancing stability, adaptability, scalability, and robustness to unlock FSL’s full potential.

\subsection{Enhancing Explainable AI}
Interpretability is critical for FSL, particularly in sensitive domains like healthcare and finance. FSL models often act as black boxes, creating trust issues. Future work can integrate Explainable AI (XAI) techniques such as \textbf{attention-based feature attribution} and \textbf{cross-modal saliency mapping} to highlight key features influencing predictions. For example, saliency maps can identify critical regions in medical images. \textbf{Counterfactual reasoning} can further enhance trust by allowing users to test “what-if” scenarios. However, balancing interpretability with computational efficiency remains a challenge, especially for resource-constrained applications.

\subsection{Leveraging Transformer Models}
Transformers, known for their self-attention mechanisms, are transformative in machine learning. Future research can adapt pre-trained transformers, like BERT and Vision Transformers (ViT), to FSL tasks. AFSL’s task-aware dynamic embeddings can enhance transformers’ generalization in low-data regimes. Reducing computational overhead is crucial for deploying these models in practical scenarios. For instance, multi-modal transformers can enhance FSL by integrating textual and visual data for applications like autonomous driving and medical diagnosis.

\subsection{Developing Scalable Frameworks}
Scalability is essential for deploying FSL on IoT devices, mobile platforms, and edge systems. Techniques like \textbf{model compression}, \textbf{quantization}, and \textbf{distributed learning} can enable FSL deployment in real-time scenarios. For example, distributed learning across decentralized nodes can enhance disaster response systems. AFSL’s modular design ensures efficient deployment while maintaining performance, making it suitable for resource-constrained environments.

\subsection{Domain-Specific Optimization}
Tailoring FSL for specific domains can significantly enhance its effectiveness. In healthcare, noise-resilient mechanisms can improve rare disease diagnosis. In climate science, hierarchical domain alignment can predict rare environmental events like wildfires. Similarly, in robotics, task decomposition and multi-modal learning enable robots to adapt to new tools or environments efficiently. Future work should develop domain-specific architectures, loss functions, and evaluation metrics to meet the unique needs of each field.

\subsection{Integrating Semi-Supervised and Unsupervised Learning}
Unlabeled data is abundant and offers opportunities to expand FSL capabilities. \textbf{Pseudo-labeling}, enhanced by AFSL’s noise-resilient module, can filter noisy labels and improve data quality. \textbf{Unsupervised meta-learning}, which creates tasks from unlabeled datasets, can further enhance FSL generalization. Combining these approaches with consistency regularization can reduce reliance on labeled datasets, particularly in underrepresented domains like rare disease research or low-resource languages.

\subsection{Expanding to Multi-Modal Learning}
The integration of multi-modal data (e.g., images, text, audio) is crucial for modern FSL applications. AFSL’s \textbf{Multi-Modal Fusion Module} demonstrates how cross-attention transformers align features across modalities. For example, autonomous driving systems can integrate sensor readings, visual data, and maps for robust decision-making. Future research should address challenges like modality-specific noise and inconsistencies while developing architectures for shared representations, enabling versatile applications in multimedia, healthcare, and robotics.

\section*{Conclusion}

Few-Shot Learning (FSL) has emerged as a transformative paradigm for addressing data scarcity, offering practical solutions for training models with minimal labeled data. While its potential is vast, limitations in interpretability, scalability, robustness, and domain adaptability remain significant barriers to broader adoption. The Adaptive Few-Shot Learning (AFSL) framework represents a substantial step forward in overcoming these challenges, enabling FSL to thrive in diverse real-world applications.

A major advancement of AFSL is in enhancing \textbf{interpretability}, a critical need in domains such as healthcare, finance, and autonomous systems. Techniques like \textbf{attention-based feature attribution}, saliency mapping, and counterfactual reasoning integrated within AFSL enhance transparency and user trust. These techniques not only facilitate user acceptance but also align with regulatory requirements in high-stakes applications, ensuring FSL models can be deployed reliably and responsibly.

The integration of \textbf{transformer-based architectures} marks another leap forward for FSL. By leveraging self-attention mechanisms, transformers optimized with AFSL’s task-aware and contextual alignment techniques enhance generalization and representation learning in low-data regimes. These innovations reduce computational overhead, making transformers suitable for FSL tasks across multi-modal learning, robotics, and natural language processing.

Addressing \textbf{scalability} is another critical milestone achieved by AFSL. Its modular design incorporates techniques like \textbf{model compression}, \textbf{quantization}, and \textbf{distributed learning}, enabling efficient deployment on resource-constrained platforms such as IoT devices, edge systems, and smartphones. Scalable FSL models extend the reach of machine learning to underserved areas like rural healthcare and real-time disaster response, unlocking transformative potential in these critical domains.

\textbf{Domain-specific optimization} further amplifies the effectiveness of FSL, tailoring models to address unique challenges in specialized fields. For instance, AFSL’s noise-resilient and domain alignment modules enhance the diagnosis of rare diseases in healthcare, prediction of extreme climate events, and adaptation of robots to novel tools and environments. These domain-focused advancements ensure that FSL solutions are fine-tuned to deliver exceptional performance in their respective applications.

The future of FSL also lies in \textbf{multi-modal learning}, where diverse data types such as images, text, and audio are seamlessly integrated. AFSL’s Multi-Modal Fusion Module leverages cross-attention transformers to align and process multi-modal data, enabling robust learning across domains like autonomous driving, multimedia analysis, and robotics. This adaptability paves the way for innovative applications requiring cross-domain transfer and integration of complex data streams.

Finally, \textbf{semi-supervised and unsupervised learning} techniques provide a pathway to further reduce reliance on labeled datasets. By integrating pseudo-labeling strategies and unsupervised meta-learning, AFSL expands training datasets and enhances generalization even in resource-constrained scenarios. These methods broaden the applicability of FSL to domains with limited labeled data while maintaining high model performance.

In conclusion, advancing Few-Shot Learning through frameworks like AFSL will make it more robust, scalable, and versatile, addressing critical challenges in interpretability, domain adaptation, and multi-modal learning. As FSL methodologies continue to evolve, they hold the potential to transform fields such as healthcare, autonomous systems, climate science, and beyond. The innovations in FSL not only promise to tackle real-world challenges but also pave the way for equitable and sustainable solutions in low-resource settings.

\end{document}